\documentclass{article}
\usepackage[preprint,sglblindworkshop]{neurips_2026}
\workshoptitle{Second Workshop on ML$\times$OR: Mathematical Foundations and Operational Integration of Machine Learning for Uncertainty-Aware Decision-Making}

\usepackage[utf8]{inputenc}
\usepackage[T1]{fontenc}
\usepackage{hyperref}
\usepackage{url}
\usepackage{booktabs}
\usepackage{amsmath, amssymb} 
\usepackage{amsthm} 
\usepackage{microtype}
\usepackage{xcolor}
\usepackage{subfigure}
\usepackage{tabularx}
\usepackage{booktabs} 
\usepackage{multirow}
\usepackage{algorithm}
\usepackage{algorithmicx}
\usepackage{algcompatible}
\usepackage{algpseudocode}
\usepackage{placeins}
\usepackage{graphicx}

\usepackage{amsmath,amsfonts,bm}

\def\eqref#1{equation~\ref{#1}}

\def\1{\bm{1}}

\DeclareMathAlphabet{\mathsfit}{\encodingdefault}{\sfdefault}{m}{sl}
\SetMathAlphabet{\mathsfit}{bold}{\encodingdefault}{\sfdefault}{bx}{n}

\DeclareMathOperator*{\argmin}{arg\,min}

\def\argmin{{\arg\min}}

\def\bp{\mathbf{p}}

\def\bbN{\mathbb{N}}

\def\bbR{\mathbb{R}}

\def\cH{\mathcal{H}}

\def\reals{{\bbR}}

\newtheorem{theorem}{Theorem}[section]

\newtheorem{lemma}[theorem]{Lemma}

\newtheorem{proposition}[theorem]{Proposition}
\newtheorem{remark}[theorem]{Remark}

\title{Improving Mixup Calibration with Wasserstein Distributionally Robust Optimization}

\author{%
  Jiaming Hu$^{1}$ \quad Yeping Jin$^{2}$ \quad Debarghya Mukherjee$^{1}$ \quad Ioannis Ch.~Paschalidis$^{2,3,4,5,6}$ \\
  $^{1}$Department of Mathematics \& Statistics \qquad $^{2}$Department of Systems Engineering \\
  $^{3}$Department of Electrical and Computer Engineering \\
  $^{4}$Division of Systems Engineering \\
  $^{5}$Department of Biomedical Engineering \qquad $^{6}$Faculty of Computing \& Data Sciences \\
  Boston University \\
  \texttt{jh7453@bu.edu \quad yepjin@bu.edu \quad mdeb@bu.edu \quad yannisp@bu.edu}
}

\begin{document}

\maketitle

\begin{abstract}

In many real-world applications, ensuring the robustness and stability of deep neural networks (DNNs) is crucial, particularly for image classification tasks that encounter various input perturbations. While Mixup-based data augmentation techniques have been widely adopted to enhance the resilience of trained models against such perturbations, our experiments reveal an important corruption robustness–calibration trade-off: stronger Mixup-based augmentation can improve robustness against corrupted data while substantially increasing expected calibration error (ECE). To address this challenge, we introduce DRO-Augment, a framework that integrates Wasserstein Distributionally Robust Optimization (W-DRO) with various Mixup-based data augmentation strategies to mitigate this trade-off. Our method substantially reduces ECE under strong Mixup-based augmentation while largely preserving corruption accuracy across CIFAR-10, CIFAR-100, CIFAR-10-C, and CIFAR-100-C. On the theoretical side, we establish novel generalization error bounds for neural networks trained using a variation-regularized loss function with augmented data, closely related to the W-DRO problem. Furthermore, we introduce a refined CIFAR-C benchmark that corrects inconsistencies in corruption intensities, providing a more reliable evaluation for future robustness research.

\end{abstract}
\section{Introduction}
Deep Neural Networks (DNNs) have become essential tools in fields such as computer vision, natural language processing, speech recognition, and autonomous systems. Their ability to model complex, non-linear relationships in large-scale datasets has led to significant progress in both research and real-world applications. By achieving state-of-the-art performance in tasks like image classification \citep{he2016deep}, speech recognition \citep{hinton2012deep,chan2016listen}, natural language processing \citep{vaswani2017attention,devlin2019bert}, DNNs have established themselves as a core component of modern artificial intelligence.

Despite their (super)-human performance on clean datasets, DNNs are often found to be highly sensitive to noisy inputs \citep{szegedy2013intriguing, gilmer2018motivating}. Various studies have demonstrated that models trained on unperturbed data can experience significant performance drops when tested on corrupted datasets \citep{rusak2020simple}, such as CIFAR-10-C and CIFAR-100-C \citep{hendrycks2019benchmarking} (which are corrupted versions of the CIFAR-10 and CIFAR-100 datasets, respectively), where typically the corruption is introduced by blurring, adding noise to the images, and in various other natural ways.
These vulnerabilities are particularly concerning in critical applications such as autonomous driving, healthcare, and security, where model failures could have serious consequences on human life or our society as a whole.


To alleviate these issues, there is a growing interest in developing models that are more robust against various natural perturbations. One key approach for achieving this is to augment the training data during training. Mixup \citep{zhang2017mixup}, Manifold Mixup \citep{verma2019manifold}, and NoisyMix \citep{erichson2022noisymix} increase training diversity through mixing examples, hidden representations, or additional perturbations, respectively. Such approaches can improve model generalization and robustness to corrupted inputs. Beyond predictive accuracy, however, reliable models should also provide well-calibrated confidence estimates \citep{guo2017calibration}. Prior work has shown that Mixup can improve calibration in a variety of settings \citep{thulasidasan2019mixup,zhang2022calibration}, while \citet{wang2023pitfall} identified limitations in the post-hoc calibratability of Mixup-trained models. More recently, \citet{qin2023effective} showed that stronger Mixup settings can induce a clean accuracy--calibration trade-off, while \citet{bouniot2025tailoring} showed that inappropriate mixing can deteriorate calibration through manifold mismatch. Our focus is complementary: we study a corruption robustness--calibration trade-off in strong Mixup-based regimes and investigate W-DRO-inspired variation regularization as a way to mitigate this trade-off while largely preserving corruption accuracy.

Another promising line of research aims to achieve robustness by optimizing a robust loss function. One of the most popular approaches in this area is Wasserstein distributionally robust optimization (W-DRO) \citep{delage2010distributionally, mohajerin2018data}, which trains a model by minimizing the worst-case expected loss over a Wasserstein ball (the radius is denoted by $\rho$) centered at the empirical distribution of the training samples:
\begin{align}
\label{eq:def_DRO}
    \min_{\beta} \sup_{\mathbb{P}: W_p(\mathbb{P}, \mathbb{Q}) \leq \rho} \mathbb{E}_{z \sim \mathbb{P}} [\ell(f_{\beta}(z))], \tag{P}
\end{align}
where \(f\) is the prediction function, \(\mathbb{Q}\) is a nominal distribution, \(\beta\) is a parameter vector to be learned, and \( \ell \in \mathcal{L} \) represents the loss function dependent on the random data \( z \). Notable studies include robust regression models in \citep{blanchet2019multivariate,chen2018robust}, adversarial training for neural networks in \citep{sinha2017certifiable}, and distributionally robust logistic regression in \citep{shafieezadeh2015distributionally, chen2021distributionally}. 
A growing body of work has shown that W-DRO effectively penalizes a certain norm of the predictor's gradient, leading to a formal connection between the standard W-DRO formulation (Equation \eqref{eq:def_DRO}) and the penalized version, known as variational regularization (see  \citep{blanchet2019robust, chen2018robust,mohajerin2018data, bartl2020robust,blanchet2019robust,blanchet2022confidence} for details). Recently, a general theory regarding the relation between this variation regularization and W-DRO has been developed in \citep{gao2024wasserstein}, which accommodates non-convex and non-smooth loss functions. W-DRO has been shown to be effective against various perturbations \citep{chen2021distributionally, bai2023wassersteindistributionalrobustnessneural, ho2023adversarial}. 

Although Mixup-based augmentation can provide strong robustness against natural corruptions, our results show that such robustness can come with a substantial calibration penalty in stronger mixing regimes. This motivates us to study whether the local stability induced by W-DRO can complement Mixup-based training by mitigating this corruption robustness--calibration trade-off.



Motivated by this observation, we propose DRO-Augment, a framework that integrates distributionally robust optimization with Mixup-based data augmentation. DRO-Augment combines augmented samples with a W-DRO-inspired gradient regularization term, promoting local stability of the loss around the training distribution. Our experiments show that DRO-Augment substantially reduces calibration error in strong-Mixup regimes while largely preserving their robustness against natural corruptions.


Our contributions can be summarized as follows: 
\vspace{-5pt}
\begin{enumerate}
    \item We empirically characterize a corruption robustness--calibration trade-off in Mixup-based training, in which mixing regimes that yield strong corruption robustness can simultaneously incur substantially worse calibration.
    \vspace{-3pt}
    \item We show that integrating W-DRO-inspired variation regularization with Mixup, Manifold Mixup, and NoisyMix improves calibration, often substantially, while largely preserving corruption robustness across CIFAR and CIFAR-C benchmarks.
    \vspace{-3pt}
    \item We establish a generalization error bound for neural networks trained using the variation-regularized W-DRO objective with augmented data, providing theoretical support for the DRO-Augment training framework.
    \vspace{-3pt}
    \item We identify inconsistencies in the original CIFAR-C severity settings and provide a refined severity calibration for more consistent robustness evaluation; details are provided in Appendix~\ref{appendix:e}.
\end{enumerate}
\vspace{-4pt}

\section{Method}
\label{sec:method}

Our DRO-Augment framework relies on two pillars: W-DRO and Mixup-based data augmentation. W-DRO aims to improve robustness against distribution shifts by guarding against worst-case distributions around the training distribution. Mixup-based data augmentation methods, on the other hand, enhance model generalization and robustness by mixing training examples or their representations. In our framework, we first apply a chosen Mixup-based augmentation method to the training data and then minimize a Wasserstein distributionally robust loss function on the augmented samples to obtain the final predictor. Our training procedure is summarized in Algorithm~\ref{alg:custom_loss_augment} in Appendix~\ref{appendix:d}. In the following, we describe the DRO-Augment procedure in detail.

At each epoch, we begin by applying data augmentation to the training minibatch to enhance data diversity. A range of Mixup-based augmentation strategies has recently gained popularity for improving robustness in prediction tasks. Our framework is flexible and can accommodate different Mixup-based augmentation techniques. In this work, we focus on three representative methods: Mixup \citep{zhang2017mixup}, a widely used technique that generates new training samples by linearly interpolating between pairs of examples and their labels; Manifold Mixup \citep{verma2019manifold}, which performs interpolation in hidden representation space; and NoisyMix \citep{erichson2022noisymix}, a state-of-the-art approach that extends feature-level mixup with noise injection and incorporates stability training \citep{zheng2016improving} to further enhance robustness. We use $\alpha_{\mathrm{mix}}$ to denote the mixing-strength hyperparameter of the corresponding Mixup-based augmentation method.

Next, rather than minimizing the standard empirical loss over the augmented data, we optimize the training objective using a W-DRO framework. 
This allows the model to be optimized against worst-case perturbations within a Wasserstein ball around the empirical distribution, providing stronger guarantees under distribution shifts. 
Given any predictor $f$, the Wasserstein distributionally robust loss $D_{P_n,\rho}(f)$ is defined as:
\[
D_{P_n,\rho}(f) = \sup_{\mathbb{P}: W_p(\mathbb{P}, \mathbb{P}_n) \leq \rho} \mathbb{E}_{(x, y) \sim P} [\ell(f(x), y)], \quad\\
\]
where \(W_p\) is the \(L_p\) Wasserstein distance and \(\mathbb{P}_n\) is the empirical distribution of $\{(X_i, Y_i)\}$. 
While this W-DRO objective provides robustness guarantees under distribution shifts, it is generally hard to optimize directly due to the inner supremum of the Wasserstein ball constraint. To address this, we adopt a variation-regularization-based approximation of the W-DRO objective, following the approach proposed by \citep{gao2024wasserstein}, which approximates the supremum with a gradient-norm-based penalty. 
This leads to the following (approximate) loss function \(R_n(f)\) of $D_{P_n, \rho}(f)$:
\begin{align}
\label{eq:reg_loss}
\textstyle
 R_n(f) = \mathbb{E}_{P_n} [\ell(f(x), y)] + \rho\left( \frac{1}{n}\sum_{i=1}^n\left\| \nabla \ell(f(x_i, y_i)) \right\|_*^q \right)^{\frac{1}{q}},
\end{align}
where \(\rho\) now serves as a penalty weight that controls the strength of the variation regularization and \(1/q + 1/p = 1\). The theoretical justification for this approximation is provided by \citet{gao2024wasserstein}; we include the formal statement and our generalization analysis in Appendix~\ref{appendix:generalization}.
This approximation makes it feasible to compute the W-DRO objective in practice. In our implementation, model parameters are updated using stochastic gradient descent (SGD) based on the regularization function \(R_n(f)\).  Over multiple epochs, the model iteratively refines its parameters to minimize both empirical risk and the regularization penalty, aiming to balance data fitting and model complexity.

\section{Experiments}
\subsection{Datasets and Metrics}
To compare the performance of our method, we conduct experiments on CIFAR-10, CIFAR-100, CIFAR-10-C, and CIFAR-100-C. CIFAR-10-C and CIFAR-100-C \citep{hendrycks2019benchmarking} contain 15 corruption types, each applied at five severity levels, and are widely used to evaluate robustness under natural corruptions. We report average classification accuracy across corruption types and severity levels as the robustness metric, together with expected calibration error (ECE) for calibration. ECE is computed using ten equal-width confidence bins of width 0.1.


\subsection{Baselines and Training Details}
\label{sec:baseline}
In this work, we focus on ResNet-based convolutional architectures, following
the convolutional setting considered by NoisyMix; extending hidden-space
mixing to Transformer token representations is beyond the scope of this study. In our experiments, we use three Mixup-based augmentation methods: Mixup, Manifold Mixup, and NoisyMix. For each method, we apply the corresponding DRO-Augment framework and evaluate performance using PreActResNet-18 \citep{he2016identity} and WideResNet-28$\times$2 \citep{zagoruyko2016wide}. Table~\ref{tab:augmentation-comparison} reports the PreActResNet-18 results, while corresponding WideResNet-28$\times$2 results and sensitivity analyses are provided in Appendix~\ref{appendix:d}. We train all models for 200 epochs and keep the hyperparameter configurations consistent across corresponding comparisons. Full training details are provided in Appendix~\ref{appendix:d}.

\subsection{Corruption Robustness--Calibration Trade-off}
We first evaluate whether DRO-Augment improves the corruption robustness--calibration profile across different Mixup-based augmentation methods. Table~\ref{tab:augmentation-comparison} compares Mixup, Manifold Mixup, and NoisyMix with their corresponding DRO-Augmented variants using PreActResNet-18.

\begingroup
\setlength{\intextsep}{0pt}
\begin{table}[H]
\centering
\caption{Average accuracy (\%) and ECE (\%) of PreActResNet-18 on CIFAR and CIFAR-C benchmarks. Higher accuracy is better and lower ECE is better.}
\small
\setlength{\tabcolsep}{1pt}
\renewcommand{\arraystretch}{1.0}
  \begin{tabular}{@{}lcccccccc@{}}
    \toprule
    \multirow{3}{*}{Method} &
      \multicolumn{4}{c}{CIFAR-10} &
      \multicolumn{4}{c}{CIFAR-100} \\
    \cmidrule(r){2-5}\cmidrule(l){6-9}
     & \multicolumn{2}{c}{Clean} & \multicolumn{2}{c}{-C} &
       \multicolumn{2}{c}{Clean} & \multicolumn{2}{c}{-C} \\
    \cmidrule(r){2-3}\cmidrule(r){4-5}\cmidrule(r){6-7}\cmidrule(l){8-9}
     & Acc. $\uparrow$ & ECE $\downarrow$ & Acc. $\uparrow$ & ECE $\downarrow$ &
       Acc. $\uparrow$ & ECE $\downarrow$ & Acc. $\uparrow$ & ECE $\downarrow$ \\
    \midrule
    Baseline        & 94.74 & 3.34 & 75.45 & 17.56 & 77.99 & 10.07 & 49.40 & 27.01 \\
    \midrule
    Mixup           & 95.60 & 14.03 & 79.25 & 12.44 & 79.12 & 14.92 & 53.29 & 11.64 \\
    Mixup + DRO     & 96.04 & 7.54 & 80.75 & 7.25 & 80.25 & 8.95 & 54.24 & 9.80 \\
    \midrule
    Manifold Mixup       & 95.83 & 17.44 & 78.71 & 15.11 & 80.19 & 12.83 & 53.31 & 12.73 \\
    Manifold Mixup + DRO & 95.86 & 9.44 & 77.99 & 10.68 & 80.56 & 9.57 & 54.35 & 11.30 \\
    \midrule
    NoisyMix        & 95.64 & 20.23 & 91.47 & 20.76 & 79.49 & 23.46 & 71.60 & 21.88 \\
    NoisyMix + DRO  & 95.73 & 8.49 & 91.72 & 7.77 & 80.44 & 15.15 & 71.46 & 12.95 \\
    \bottomrule
  \end{tabular}
\label{tab:augmentation-comparison}
\end{table}
\endgroup

Across all three augmentation methods, adding DRO consistently reduces ECE while largely preserving corruption accuracy. On CIFAR-10-C and CIFAR-100-C, ECE decreases in all six method--dataset comparisons. Corruption accuracy changes by at most 1.50 percentage points across these six comparisons and improves in four of the six cases, indicating that the calibration gains are obtained with little change in corruption robustness. The largest calibration improvements occur for NoisyMix: ECE decreases from 20.76\% to 7.77\% on CIFAR-10-C and from 21.88\% to 12.95\% on CIFAR-100-C, while corruption accuracy changes by only +0.25 and -0.14 percentage points, respectively. A similar qualitative pattern is observed with WideResNet-28$\times$2, where ECE also decreases in all six method--dataset comparisons and corruption accuracy changes by at most 0.99 percentage points; see Appendix~\ref{appendix:d}.

\vspace{-0.6em}
To examine the corruption robustness--calibration trade-off as a function of mixing strength, we additionally vary the mixing-strength parameter $\alpha_{\mathrm{mix}}$ in Appendix~\ref{appendix:d}. These sensitivity results show that stronger mixing regimes can incur substantially larger calibration error, although the relationship between $\alpha_{\mathrm{mix}}$ and corruption accuracy is not strictly monotonic.
\vspace{-0.8em}


\section{Conclusion and Discussion}
In this work, we identify a corruption robustness--calibration trade-off in Mixup-based training: regimes that yield strong corruption robustness can incur substantially worse calibration. DRO-Augment combines Mixup-based augmentation with W-DRO-inspired variation regularization and substantially reduces ECE while largely preserving corruption robustness across Mixup, Manifold Mixup, and NoisyMix. Our theoretical analysis provides a generalization bound for neural networks trained with the variation-regularized objective and augmented data. We also provide a refined CIFAR-C severity calibration for more consistent robustness evaluation. DRO-Augment introduces additional computation from the gradient regularization term; future work will investigate whether similar corruption robustness--calibration trade-offs and the benefits of variation regularization extend to broader architectures and learning settings.

\bibliography{ref}
\bibliographystyle{plainnat}

\appendix
\section{Notations}
Scalars and individual data points are denoted by lowercase letters (e.g., $x$, $y$). The input-output pair is denoted by $z = (x, y)$. Prediction functions are written as $f(x)$, with associated loss $\ell(f(x), y)$ and input gradient $\nabla \ell(f(x), y)$. The true data distribution is $P_{\text{true}}$ and the empirical distribution over $n$ samples is $P_n$.
Calligraphic letters (e.g., $\mathcal{H}^\alpha$, $\mathcal{F}$) denote function classes or sets.
We use $\|\cdot\|_{q^*}$ for the dual norm of $\ell_q$. 

\section{Auxiliary Lemmas}

\begin{lemma}[\citep{gao2024wasserstein}]
\label{lem:1}
Under the assumptions that the data distribution satisfies a bounded density condition around the set of all nondifferentiable points of \(f: D_f\), there are \(\bar{\rho},J > 0\) such that for all \(\rho < \bar{\rho}\), with probability at least \(1 - e^{-t}\), for every \(f\) from a hypothesis family \(\mathcal{F}\):
\begin{align*}
D_{P_n,\rho}(f) \ = \ &\min_{f \in \mathcal{F}} \Biggl\{ \frac{1}{n} \sum_{i=1}^n \ell\bigl(f(x_i),\,y_i\bigr)\;+\;\rho \,\mathbb{E}_{\mathbb{P}_n} \Bigl[ \ell'\bigl(f(x), y\bigr)\,\bigl\lVert \nabla f(x) \bigr\rVert_* \Bigr]\Biggr\} \\
&+ \rho\,\sqrt{\frac{t}{2n}} + \rho^2 (J + ||H||_{\mathbb{P}_{n,1}}) + 2\rho\,\mathbb{E}\bigl[\mathcal{R}_n(\mathcal{J}_\rho)\bigr]
\end{align*}
where \(\mathcal{J}_{\rho,  \mathcal{F}}:= \left\{x \mapsto {1}\{d(x, D_f) < \rho\} : f \in \mathcal{F}, \, D_f \neq \emptyset \right\}\) and \(H(x)\) is an upper bound on the operator norm of the Hessian of \(f(x)\).
\end{lemma}

\begin{lemma}[\citep{zhang2020does}]
\label{lem:2}
Suppose the function class \(\mathcal{F}\) is defined over \(\mathcal{X}\) and satisfies \(\sup_{f \in \mathcal{F}} \|f\|_\infty \leq D\). For any samples \(x_1, \ldots, x_n\) from \(\mathcal{X}\), we have

\begin{align*}
\mathbb{E}_\sigma \left[
\sup_{f \in \mathcal{F}} \frac{1}{n} \sum_{i=1}^n \sigma_i f(x_i)
\right]
\leq &\inf_{\delta \geq 0} \bigg\{
    4\delta +
    12 \int_\delta^D \sqrt{
        \frac{\log \mathcal{N}(u, \mathcal{F}, L_2(P_n))}{n}
    } \, du
\bigg\}
\end{align*}

where \(\sigma = (\sigma_1, \ldots, \sigma_n)\) are i.i.d. Rademacher variables and \(L_2(P_n)\) denotes the data-dependent \(L_2\) metric.
\end{lemma}

\begin{lemma}[\citep{belomestny2023simultaneous}, Theorem 1]
\label{lem:3}
Fix \(\alpha > 2\) and \(p,d \in \mathbb{N}\). Then, for any function \(f : [0,1]^d \rightarrow \mathbb{R}^p\) with \(f \in \mathcal{H}^\alpha([0,1]^d)\), for any integer \(Z \geq 2\), there exists a neural network \(h_f\) with ReQU activations such that it has \(\mathcal{O}(\log d + \lfloor \alpha \rfloor )\) layers, at most \(\mathcal{O}(p \vee d(Z + \lfloor \alpha \rfloor)^d)\) neurons per layer, and \(\mathcal{O}(p(d\alpha + d^2 (Z + \lfloor \alpha \rfloor)^d)\) nonzero weights in \([-1,1]\), satisfying
\[
\| f - h_f \|_{\mathcal{H}^\ell([0,1]^d)} \leq \frac{C ^{\alpha d} \alpha^\ell}{Z^{\alpha - \ell}} \quad \text{for all } \ell \in \{0, \ldots, \lfloor \alpha \rfloor\}.
\]
\end{lemma}

\begin{lemma}[\citep{bartlett1998almost}, Theorem 2.1]
\label{lem:4}
For any positive integers \(U\), \(k \leq U\), \(L \leq U\), \(l\), and \(p\), considering a network with real inputs, up to \(U\) parameters, \(k\) computational units in \(L\) layers, a single output unit (identity activation), and all other units with piecewise polynomial activation of degree \(l\) and \(p\) breakpoints, the VC-dimension satisfies
\[
\text{VCdim}(\text{sgn}(\mathcal{F})) \leq 2UL \log(2eULpk) + 2UL^2 \log(l+1) + 2L.
\]
Furthermore, since \(L, k = \mathcal{O}(U)\), for fixed \(l\) and \(p\):
\[
\text{VCdim}(\text{sgn}(\mathcal{F})) = \mathcal{O}(UL \log U + UL^2).
\]
\end{lemma}

\begin{lemma}[\citep{anthony2009neural}, Theorem 12.2]
\label{lem:5}
Let \(\mathcal{F}\) be a set of real functions from a domain \(\mathcal{X}\) to \([0, M]\). Let \(\varepsilon > 0\) and \(\mathrm{Pdim}(\mathcal{F})\) denote the pseudo-dimension. If \(n \geq \mathrm{Pdim}(\mathcal{F})\), then the uniform covering number satisfies
\[
\mathcal{N}_\infty(\varepsilon, \mathcal{F}, n) \leq \left( \frac{e n M}{\varepsilon \, \mathrm{Pdim}(\mathcal{F})} \right)^{\mathrm{Pdim}(\mathcal{F})}.
\]
\end{lemma}

\begin{lemma}[\citep{vershynin2018high}, Corollary 4.2.13]
\label{lem:6}
The covering numbers of the unit Euclidean ball \(B_2^d\) satisfy for any \(\varepsilon \in (0,1]\):
\[
\left( \frac{1}{\varepsilon} \right)^d \leq \mathcal{N}\left( B_2^d, \varepsilon \right) \leq \left( \frac{3}{\varepsilon} \right)^d.
\]
\end{lemma}

\section{Generalization Analysis}
\label{appendix:generalization}
\subsection{Variation-Regularized Approximation of W-DRO}
\begin{proposition}[Theorem 1 in \citep{gao2024wasserstein}]
\label{prop 2.1}
Let \( z = (x,y) \in \mathcal{Z} = \mathcal{X} \times \mathcal{Y}\). If the gradient and Hessian norms of \(f\) are well bounded, for many important applications, it can be shown that when \(\rho =O(1/\sqrt{n})\),  the following asymptotic equation holds:

$\min_{f}\; D_{P_n,\rho}(f)
  = \min_{f}\Bigl\{
      \mathbb{E}_{P_n}\!\left[\ell\bigl(f(x,y)\bigr)\right]   +\rho\Bigl(
        \tfrac{1}{n}\sum_{i=1}^{n}
    \lVert\nabla\ell\bigl(f(x_i,y_i)\bigr)\rVert_{*}^{q}
      \Bigr)^{\!\!1/q}
      + O(n^{-1})
     \Bigr\}.
$
\end{proposition}
\subsection{Generalization Error Bound for Neural Networks}
In this section, we present the asymptotic generalization error bound for a neural-network-based estimator obtained via optimizing the variation-regularized loss function $R_n(f)$ (Equation \eqref{eq:reg_loss}) with augmented data. 
As illustrated in Section \ref{sec:method}, $R_n(f)$ serves as a proxy of the W-DRO and the approximation error depends on $\rho$. For our theoretical analysis, we focus on the variation-regularization-based approximation of the $L_\infty$-Wasserstein DRO optimization problem, i.e., in our case,
\begin{equation*}
\textstyle
R_n(f) = \frac1n \left\{\sum_i \ell(f(x_i), y_i) + \rho \|\nabla\ell(f(x_i), y_i)\| \right\}
\end{equation*}

Generalization error bounds for neural networks are crucial for understanding how well a trained model will perform on unseen data, providing theoretical guarantees that guide reliable deployment in practice. 
In recent years, there has been a surge of research focused on generalization error bounds for structured neural networks in nonparametric regression and classification settings (e.g., see \cite{kohler2021rate, schmidt2020nonparametric, bhattacharya2024deep} and references therein). 
To set up notations, given any $L \in \bbN$ and a vector $\bp = (p_0, p_1, \cdots, p_{L+1})$, a neural network with depth $L$ and width vector $\bp$ is defined as: 
\[
f(x) = W_L \circ \sigma \circ W_{L-1} \circ \sigma \circ \cdots \circ \sigma \circ W_0 x,
\]
where $W_i \in \mathbb{R}^{p_{i+1} \times p_{i}}$, and each layer includes corresponding bias vectors \( v_h \in \mathbb{R}^{p_{h+1}} \). The activation function $\sigma$ is applied to each component of the input vector, which is taken to be $\sigma(x) = (\max\{0, x\})^2$ in our analysis.  In practice, our estimator is trained not only with the W-DRO but also with
augmented samples generated by mixup. For any two sample $(X_i, Y_i), (X_j, Y_j)$ (generated from a distribution, namely $P_{\mathrm{true}}$), and $\lambda_{ij} \sim \mathrm{Beta}(\alpha,\beta)$, we define their mixture as $Z_{ij} = \lambda_{ij}(X_i, Y_i) + (1-\lambda_{ij})(X_j, Y_j)$. The population distribution of $Z_{ij}$ is denoted as $P_{\rm mix}$. 
To preserve independence within the mixed sample, we construct the mixup dataset using disjoint pairs. Specifically, with a slight abuse of notation, we define 
$
Z_{i} \;=\; \lambda_i \,(X_{2i-1}, Y_{2i-1}) \;+\; (1 - \lambda_i)\,(X_{2i}, Y_{2i})
$, for $1 \le i \le n/2$, where we assume $n$ is even; if $n$ is odd, the last sample is discarded.
Let $R_n^{mix}(f)$ denote the empirical risk (counterpart of $R_n(f)$) computed using the mixup samples $\{Z_i\}_{1 \le i \le n/2}$. Our estimator $\hat f^{mix}_{NN}$ is defined as: 
\begingroup
\small
\begin{equation}
\label{eq:nn_min_reg}
\textstyle
\hat f^{mix}_{NN} = \argmin_{f \in \mathcal{NN}_{U, L, \bp}^{a_1, a_2}} \ R_n^{mix}(f) \,.
\end{equation}
\endgroup
where $\mathcal{NN}_{U, L, \bp}^{a_1, a_2}$ is the collection of all neural networks with depth $L$, width vector $\bp$, total number of active weights $U = \sum_{h = 0}^{L}(\|W_h\|_0 + \|v_h\|_0)$, norm of gradient is almost $a_1/2$ and (operator) norm of the Hessian is almost $a_2/2$. The robust generalization error of $f$ with respect $L_\infty$ W-DRO loss is defined as: 
\begingroup
\small
\begin{align*}
\textstyle
D_{P_{\rm true}, \rho}(f)
& = \mathop{\sup}_{Q:\, W_\infty(P_{\rm true}, Q) \le \rho}
    \, \mathbb{E}_{(x, y) \sim P_{\rm true}}[\ell(f(x), y)] \\
& = \mathbb{E}_{(x, y) \sim P_{\rm true}} \left[
    \sup_{\tilde x:\, \|x - \tilde x\|_2 \le \rho}
    \ell(f(\tilde x), y)
\right]
\end{align*}
\endgroup
We define $D_{P_{\mathrm{mix}}, \rho}(f)$ as the robust generalization error of $f$ with respect to $P_{\rm mix}$,  
and define $f_*^{mix}$ to be the minimizer of $D_{P_{\mathrm{mix}}, \rho}(f)$ over class of H\"{o}lder function $\cH^{\alpha}(\reals^d)$, defined as follows: 
\begingroup
\small
\begin{align*}
\mathcal{H}^\alpha(\mathbb{R}^d)
= \bigg\{ f : \mathbb{R}^d \to 
\mathbb{R} \,\bigg|\, 
 \max_{\|\mathbf{s}\|_1 \leq r} \sup_{x \in \mathbb{R}^d}
  \left| \partial^{\mathbf{s}} f(x) \right| \leq 1, 
 \max_{\|\mathbf{s}\|_1 = r} \sup_{x_1 \neq x_2}
  \frac{ \left| \partial^{\mathbf{s}} f(x_1) - \partial^{\mathbf{s}} f(x_2) \right| }
       { \|x_1 - x_2\|_\infty^\beta } \leq 1
\bigg\}
\end{align*}

\endgroup
The goal of the generalization bound is to provide an upper bound on the difference between $D_{P_{\mathrm{mix}}, \rho}(\hat f^{mix}_{ NN})$ and $D_{P_{\mathrm{mix}}, \rho}(f_*^{mix})$, which we present in the following theorem: 
\begin{theorem}
\label{theorem 5.1}
Assume that the population minimizer $f_*^{mix} \in \cH^{\alpha}(\reals^d)$ with $\alpha > d/2$. 
For any fixed $Z > 0$, if the network has \(O(\log d + \left\lfloor \alpha \right\rfloor )\) layers, with the \(
\max_{i} p_i
 = O(p_{L+1} \vee d(Z+ \left\lfloor \alpha \right\rfloor)^d)\) and \(O(p_{L+1}d(d + \alpha )(Z+ \left\lfloor \alpha \right\rfloor)^d)\) non-zero weights taking their values in [-1,1], then the optimizer $\hat f^{mix}_{NN}$, as defined in Equation \eqref{eq:nn_min_reg}, with $\ell$-Lipschitz loss function, satisfies that there exists $c, \bar{\rho} > 0  \, \text{such that for all} \,\rho < \bar{\rho}$, with probability at least $1 - n^{-c}$, 
\begin{align*}
\big| D _{P_{\text{mix}},\rho}(\hat f^{mix}_{NN}) - D _{P_{\text{mix}},\rho}(f_*^{mix} ) \big|
\leq  C_1 \bigg(
  \sqrt{\frac{\log n \cdot (U + \log U)}{n}} 
  + U^{-\alpha/d} 
  + \rho \sqrt{\frac{\log n}{n}} 
  + \rho^2 
\bigg) \,.
\end{align*}
If we further select  \( U \asymp (n/\log{n})^\frac{d}{2\alpha + d} \),   then we have
\begin{equation*}
\textstyle
\big| D _{P_{\text{mix}},\rho}(\hat f^{mix}_{NN}) - D _{P_{\text{mix}},\rho}(f^{mix} _*) \big|
\leq 
  C_2 \bigg(
  \left( \frac{\log n}{n} \right)^{\frac{\alpha}{2\alpha + d}}  + \rho \sqrt{\frac{\log n}{n}} 
  + \rho^2
\bigg) \,.
\end{equation*}
\textit{where \(c_1, c_2, C_1, C_2 \) are fixed constants, depending on $(\alpha, d, a_1, a_2)$.}
\end{theorem}

\begin{remark}
Although recently \cite{liu2024nonasymptotic} established a similar bound on a neural network-based estimator under $L_\infty$ perturbation, our analysis differs from theirs in various aspects; first, we obtain $\hat f^{mix}_{NN}$ by minimizing the variation approximated loss with augmented data $R_n^{mix}(f)$, whereas they did not consider the augmentation nor the variational approximation. 
Secondly, the class of neural networks considered in our theoretical analysis differs from them: we focus on sparsely connected networks, where sparsity is imposed through conditions on $U$, which can be implemented via dropout in practice, in contrast to the fully connected networks studied in their paper. 
Additionally, we employ the ReQU activation function, rather than the ReLU activation function used in their paper. As a result of these differences, we are able to achieve a faster rate of convergence: our estimator attains a convergence rate of $n^{-\frac{\alpha}{2\alpha + d}}$, compared to their rate $n^{-\frac{\alpha}{3\alpha + 2d}}$. A full derivation of the theoretical bound is provided in Appendix C, where we also include the approximation formulation for the DRO-Augment objective.
\end{remark}

\begin{remark}
   In our theoretical development, we assume that mixing is performed on independent pairs, ensuring that the resulting mixed samples remain independent. In practice, however, one might consider forming all possible pairs, which would introduce correlations among the mixed samples. Our framework can be extended to this setting using empirical process techniques for U-statistics. Nevertheless, such an extension offers no additional insight beyond more cumbersome mathematical bookkeeping, and we therefore omit it here.
\end{remark}

\subsection{Risk Bound Proof}
The following proof adapts the approach from \citep{liu2024nonasymptotic} with modifications to account for W-DRO, augmented samples and the ReQu activation function. Throughout the proof below, let
n denote the size of the augmented dataset.\\
The decomposition of the risk function is given by:
\begin{align*}
& D _{P_{\mathrm{mix}},\rho}(\hat f^{mix}_{NN}) -   D _{P_{\mathrm{mix}},\rho}(f_*^{mix})\\ 
&= 
  D _{P_{\mathrm{mix}},\rho}(\hat f^{mix}_{NN}) -  D^{mix}_{P_{n},\rho}(\hat f^{mix}_{NN}) + D^{mix}_{P_{n},\rho}(\hat f^{mix}_{NN}) - R_n^{mix}(\hat f^{mix}_{NN})  
 + R_n^{mix}(\hat f^{mix}_{NN}) \\
 &- R_n^{mix}(\bar{f}) 
 + R_n^{mix}(\bar{f}) - D^{mix}_{P_{n},\rho}(\bar{f}) 
+  D^{mix}_{P_{n},\rho}(\bar{f}) - D^{mix}_{P_{n},\rho}(f_*^{mix})
+ D^{mix}_{P_{n},\rho}(f_*^{mix}) -  D _{P_{\mathrm{mix}},\rho}(f_*^{mix})\\
&\leq  D _{P_{\mathrm{mix}},\rho}(\hat f^{mix}_{NN}) -  D^{mix}_{P_{n},\rho}(\hat f^{mix}_{NN}) + D^{mix}_{P_{n},\rho}(\hat f^{mix}_{NN}) - R_n^{mix}(\hat f^{mix}_{NN})  
+ R_n^{mix}(\bar{f})\\
&- D^{mix}_{P_{n},\rho}(\bar{f}) 
+  D^{mix}_{P_{n},\rho}(\bar{f}) - D^{mix}_{P_{n},\rho}(f_*^{mix})+ D^{mix}_{P_{n},\rho}(f_*^{mix}) -  D _{P_{\mathrm{mix}},\rho}(f_*^{mix})\\
\end{align*}
where $D^{mix}_{P_{n}, \rho}(f)$ is the empirical counterpart of $D_{P_{n}, \rho}(f)$ computed on mixup samples. Therefore, 
\begin{align*}
| D _{P_{\mathrm{mix}},\rho}(\hat f^{mix}_{NN}) -   D _{P_{\mathrm{mix}},\rho}(f_*^{mix}) | 
& \leq  | D _{P_{\mathrm{mix}},\rho}(\hat f^{mix}_{NN}) -  D^{mix}_{P_{n},\rho}(\hat f^{mix}_{NN})| \\
&+ |D^{mix}_{P_{n},\rho}(\hat f^{mix}_{NN}) - R_n^{mix}(\hat f^{mix}_{NN})| \\
&+ |R_n^{mix}(\bar{f}) - D^{mix}_{P_{n},\rho}(\bar{f})| \\
& + |D^{mix}_{P_{n},\rho}(\bar{f}) - D^{mix}_{P_{n},\rho}(f_*^{mix})|\\
& + |D^{mix}_{P_{n},\rho}(f_*^{mix}) -  D _{P_{\mathrm{mix}},\rho}(f_*^{mix})|
\end{align*}
where $\bar{f} \in \mathcal{NN}_{U, L}^{a_1, a_2} $ is the approximation of $f_*^{mix}$, then we have 
\[
\begin{cases}
 D _{P_{\mathrm{mix}},\rho}(\hat f^{mix}_{NN}) -  D^{mix}_{P_{n},\rho}(\hat f^{mix}_{NN}) = B_1,\\

D^{mix}_{P_{n},\rho}(\hat f^{mix}_{NN}) - R_n^{mix}(\hat f^{mix}_{NN}) = B_2,\\

R_n^{mix}(\bar{f}) - D^{mix}_{P_{n},\rho}(\bar{f}) = B_3,\\

D^{mix}_{P_{n},\rho}(\bar{f}) - D^{mix}_{P_{n},\rho}(f_*^{mix}) = B_4,\\
D^{mix}_{P_{n},\rho}(f_*^{mix}) -  D _{P_{\mathrm{mix}},\rho}(f_*^{mix}) = B_5.
\end{cases}
\]

\subsubsection{Bound for \(B_2\) and \(B_3\)} 

By Lemma~\ref{lem:1}, we directly derive that for any function \(f \in \mathcal{NN}_{U, L}^{a_1, a_2} \) with the ReQU activation function, the following holds with probability at least \(1 - e^{-t}\), 
\begin{align*}
|R_n^{mix}(f) - D^{mix}_{P_n,\rho}(f)| =\rho\,\sqrt{\frac{t}{2n}}
+ \rho^2 (J + \frac{a_2}{2})
\end{align*}
Hence we can get the asymptotic bound of \(B_2\) and \(B_3\).

\subsubsection{Bound for \(B_4\)}

Define the approximation error by
\[
\mathcal{E}(\mathcal{H}^\alpha, \mathcal{NN}_{U, L}^{a_1, a_2} ) = \| f - \bar{f} \|_{\mathcal{H}^0([0,1]^d)}
\]

There exists \(\bar{f} \in \mathcal{NN}_{U, L}^{a_1, a_2} \) approximating the target function \(f_*^{mix} \in \mathcal{H}^\alpha\) such that
\[
\| f_*^{mix} - \bar{f} \|_{\mathcal{H}^0([0,1]^d)} = O\Bigl(\mathcal{E}(\mathcal{H}^\alpha, \mathcal{NN}_{U, L}^{a_1, a_2} )\Bigr).
\]
Indexing augmented samples by $t$ via \(Z_t= (\tilde{X_{t}},\tilde{Y_{t}})\in{P}_{\mathrm{mix}} \ \text{s.t.}\
 \tilde{X_{t}}=\lambda_t X_{2t}+(1-\lambda_t)X_{2t -1},\ \tilde{Y_{t}}=\lambda_t Y_{2t}+(1-\lambda_t)Y_{2t -1},\ \text{where} \ \lambda_{t}\stackrel{\text{i.i.d.}}{\sim}\mathrm{Beta}(\alpha,\beta).
\)
Since \(\ell\) is \(L_\ell\)-Lipschitz,  the difference between \(D^{mix}_{P_n,\rho}(f_*^{mix})\) and \(D^{mix}_{P_n,\rho}(\bar{f})\) satisfies
\begin{align*}
D^{mix}_{P_n,\rho}(\bar{f}) - D^{mix}_{P_n,\rho}(f_*^{mix})
\leq &\frac{1}{n} \sum_{t=1}^n \Biggl|\sup_{\|\tilde{X'_{t}} - \tilde{X_{t}}\|\leq \rho} 
      \ell\bigl(\bar{f}(\tilde{X'_{t}}), \tilde{Y_{t}}\bigr)
 - \sup_{\|\tilde{X'_{t}} - \tilde{X_{t}}\|\leq \rho} \ell\bigl(f_*^{mix}(\tilde{X'_{t}}), \tilde{Y_{t}}\bigr) \Biggr|
\\
=&\frac{1}{n}  \sum_{t=1}^n \Biggl|\sup_{\|\tilde{X'_{t}} - \tilde{X_{t}}\|\leq \rho} 
      \ell\bigl(f_*^{mix}(\tilde{X'_{t}}), \tilde{Y_{t}}\bigr)
 - \sup_{\|\tilde{X'_{t}} - \tilde{X_{t}}\|\leq \rho} \ell\bigl(\bar{f}(\tilde{X'_{t}}), \tilde{Y_{t}}\bigr) \Biggr|\\
\leq& \frac{1}{n} \sum_{t=1}^n \sup_{\|\tilde{X'_{t}} - \tilde{X_{t}}\|\leq \rho}
\Bigl|\ell\bigl(f_*^{mix}(\tilde{X'_{t}}), \tilde{Y_{t}}\bigr)
- \ell\bigl(\bar{f}(\tilde{X'_{t}}), \tilde{Y_{t}}\bigr)\Bigr|\\
\leq& L_\ell\,\|f_*^{mix} - \bar{f}\|_{\mathcal{H}^0([0,1]^d)} 
\end{align*}
Based on Lemma~\ref{lem:3}, we derive
\[
B_4 \leq \frac{L_\ell C ^{\alpha d}}{Z^{\alpha} }
\]

\subsubsection{Bound for \(B_5\)}

For any \(f \in \mathcal{H}^\alpha\) and \(z = (x,y) \in \mathcal{Z}\), define
\[
\tilde{\ell}(f,z) = \sup_{\|x'-x\| \leq \rho} \ell\bigl(f(x'),y\bigr).
\]

We have
\begin{align*}
B_5 = D^{mix}_{P_n,\rho}(f_*^{mix}) -  D _{P_{\mathrm{mix}},\rho}(f_*^{mix}) 
 \leq \sup_{f \in \mathcal{NN}_{U, L}^{a_1, a_2} } \Bigl\{ &\mathbb{E}_{P_n,\rho}\bigl[\tilde{\ell}(f,Z_{t})\bigr]
 - \mathbb{E}_P\bigl[\tilde{\ell}(f,Z_{t})\bigr] \Bigr\}.
\end{align*}\\
Define the class 
\[
\mathcal{L}^\alpha = \Bigl\{ \tilde{\ell}(f, \cdot) : \mathcal{Z} \to \mathbb{R} \mid f \in \mathcal{H}^\alpha \Bigr\},
\]
and let the random vector \(\sigma = (\sigma_1, \ldots, \sigma_n)\) consist of i.i.d. Rademacher variables that are independent of the data. Denote the samples by 
\(
Z_{1:n} = \{Z_{t}\}_{t=1}^n,\;\text{with } Z_{t}=(\tilde{X_{t}},\tilde{Y_{t}}).
\)
Let 
\[
Z_{t}' = (\tilde{X_{t}}',\tilde{Y_{t}}'),\quad t = 1, \ldots, n,
\]
be drawn i.i.d. from \(P_{\mathrm{mix}}\) (the ghost sample). Then,
\[D^{mix}_{P_n,\rho}(f_*^{mix}) -  D _{P_{\mathrm{mix}},\rho}(f_*^{mix})\]
\[
\begin{aligned}
&\le \sup_{f \in \mathcal{NN}_{U, L}^{a_1, a_2} } \mathbb{E}_\sigma \Biggl\{ \mathbb{E}_{P_n,\rho}\bigl[\tilde{\ell}(f,Z_{t})\bigr] - \mathbb{E}_P\bigl[\tilde{\ell}(f,Z_{t})\bigr] \Biggr\} \\
&= \sup_{f \in \mathcal{NN}_{U, L}^{a_1, a_2} } \mathbb{E}_\sigma \Biggl\{ \frac{1}{n} \sum_{t=1}^n \tilde{\ell}(f,Z_{t}) - \mathbb{E}_{Z_{1:n}'}\Bigl[\frac{1}{n} \sum_{t=1}^n \tilde{\ell}(f,Z_{t}')\Bigr] \Biggr\} \\
&\le \mathbb{E}_{Z_{1:n}'} \mathbb{E}_\sigma \Biggl\{ \sup_{f \in \mathcal{NN}_{U, L}^{a_1, a_2} } \Bigl[ \frac{1}{n} \sum_{t=1}^n \tilde{\ell}(f,Z_{t}) - \tilde{\ell}(f,Z_{t}') \Bigr] \Biggr\} \\
&= \mathbb{E}_\sigma \Biggl\{ \sup_{f \in \mathcal{NN}_{U, L}^{a_1, a_2} } \frac{1}{n} \sum_{t=1}^n \sigma_t \Bigl[\tilde{\ell}(f,Z_{t}) - \tilde{\ell}(f,Z_{t}') \Bigr] \Biggr\} \\
&=\mathcal{R}_n(\mathcal{L}^\alpha).
\end{aligned}
\]
Therefore, $B_5 \leq {\mathcal{R}}_n(\mathcal{L}^\alpha)$.
Because for any $f, \tilde{f} \in \mathcal{H}^\alpha$ satisfying $\|f - \tilde{f}\|_\infty \leq u / \text{Lip}^1(\ell)$, it follows
\[
\left|\tilde{\ell}(f,Z_{t}) - \tilde{\ell}(\tilde{f}, Z_{t})\right| = \left|\sup_{\tilde{X_{t}}' \in B_\epsilon(\tilde{X_{t}})} \ell(f(\tilde{X_{t}}'), \tilde{Y_{t}}) - \sup_{\tilde{X_{t}}' \in B_\epsilon(\tilde{X_{t}})} \ell(\tilde{f}(\tilde{X_{t}}'), \tilde{Y_{t}})\right|
\]
\[
\leq \text{Lip}^1(\ell) \sup_{\tilde{X_{t}}' \in B_\epsilon(\tilde{X_{t}})} \left|f(\tilde{X_{t}}') - \tilde{f}(\tilde{X_{t}}')\right|
\leq u.
\]
Hence, we have
\[
\log \mathcal{N}(u, \mathcal{L}^\alpha, \|\cdot\|_\infty) \leq \log \mathcal{N}\left(\frac{u}{\text{Lip}^1(\ell)}, \mathcal{H}^\alpha, \|\cdot\|_\infty\right).
\]
Based on the Kolmogorov–Tikhomirov result (1959),
\[
\log \mathcal{N}\Bigl(u, \mathcal{L}^\alpha, \|\cdot\|_\infty\Bigr) \lesssim u^{-d/\alpha}.
\]
Let \(L_2(P_n)\) denote the \(L_2\) metric induced by the samples. Then,
\[
\log \mathcal{N}\Bigl(u, \mathcal{L}^\alpha, L_2(P_n)\Bigr) \le \log \mathcal{N}\Bigl(u, \mathcal{L}^\alpha, \|\cdot\|_\infty\Bigr) \lesssim u^{-d/\alpha}.
\]
Since \(\sup_{f \in \mathcal{H}^\alpha}\|f\|_\infty \le 1\), there exists a constant \(B\) such that
\[
\sup_{Z_{t} \in {P}_{\mathrm{mix}}} |\tilde{\ell}(f,Z_{t})| \le B \quad \text{for any } f \in \mathcal{H}^\alpha.
\]
Combining this with Lemma~\ref{lem:2} and the arguments in \citep{liu2024nonasymptotic} (Section C.2.3), we get:
\[
\hat{\mathcal{R}}_n(\mathcal{L}^\alpha) \lesssim \inf_{\delta \geq 0} \left\{ 4\delta + 12 \int_\delta^1 \sqrt{\frac{\log \mathcal{N}(u, \mathcal{L}^\alpha, \|\cdot\|_\infty)}{n}} \, du \right\}
\]
\[
\lesssim \inf_{\delta \geq 0} \left\{ \delta + n^{-1/2} \int_\delta^1 u^{-d/(2\alpha)} \, du\right\}
\]
\[
\lesssim n^{-\min\{1/2, \alpha/d\}} \log^{c(\alpha, d)} n.
\]
Therefore,
\[
B_5 = D^{mix}_{P_n,\rho}(f_*^{mix}) -  D _{P_{\mathrm{mix}},\rho}(f_*^{mix}) \lesssim n^{-\min\{1/2, \alpha/d\}} \log^{c(\alpha, d)} n.
\]

\subsubsection{Bound for \(B_1\)}
Define the class of functions \(\mathcal{L}_n\) by
\[
\mathcal{L}_n = \left\{ \tilde{\ell}(f, z) : \mathcal{Z} \mapsto \mathbb{R} \mid f \in \mathcal{NN}_{U, L}^{a_1, a_2}  \right\}.
\]
For a given set of samples \(z_1, \ldots, z_n\) from \(\mathcal{Z}\), the empirical Rademacher complexity of class \(\mathcal{L}_n\) is defined by
\[
\widehat{\mathcal{R}}_n(\mathcal{L}_n) = \mathbb{E}_\sigma \left\{ \sup_{f \in \mathcal{NN}_{U, L}^{a_1, a_2} } \frac{1}{n} \sum_{i=1}^n \sigma_i \tilde{\ell}(f, z_i) \right\}.
\]
As the similar logic for \(B_5\), we could see that 
\[
| D _{P_{\mathrm{mix}},\rho}(\hat f^{mix}_{NN}) -  D^{mix}_{P_{n},\rho}(\hat f^{mix}_{NN})| \leq \widehat{\mathcal{R}}_n(\mathcal{L}_n)
\]
where \(\hat f^{mix}_{NN} \in \mathcal{NN}_{U, L}^{a_1, a_2} \). For a given \( \tau \in (0, \rho) \), let \( C_{B_\rho}(\tau) \) be a \((\tau, \|\cdot\|_\infty)\)-cover of \( B_\rho(0) \) with the smallest cardinality \(M_\tau\). Denote the elements of \(C_{B_\rho}(\tau)\) by \(\delta_1, \ldots, \delta_{M_\tau}\). It follows by Lemma~\ref{lem:6} that
\[
\log M_\tau \leq cd \log(\rho \tau^{-1})
\]
for a constant \(c\).
 For any \(Z_{t} = (\tilde{X_{t}}, \tilde{Y_{t}}) \in {P}_{\mathrm{mix}}\), the continuity of \(\ell\) and \(f\) imply that there exists \(\delta' \in B_\rho(0)\) such that 
\begin{align*}
\tilde{\ell}(f, Z_{t}) &= \ell(f(\tilde{X_{t}} + \delta'), \tilde{Y_{t}}) - \max_k \ell(f(\tilde{X_{t}} + \delta_k), \tilde{Y_{t}})\\
&\leq \min_k |\ell(f(\tilde{X_{t}} + \delta'), \tilde{Y_{t}}) - \ell(f(\tilde{X_{t}} + \delta_k), \tilde{Y_{t}})|\\ &\leq \text{Lip}^1(\ell) \text{Lip}(f) \|\delta' - \delta_k\|_\infty \\ &\leq \text{Lip}^1(\ell) \text{Lip}(f) \tau.
\end{align*}

Therefore, for any \(f \in \mathcal{NN}_{U, L}^{a_1, a_2} \),

\begin{align*}
\frac{1}{n} \sum_{t=1}^n \sigma_t \tilde{\ell}(f, Z_{t})
= &\frac{1}{n} \sum_{t=1}^n \bigg[
    \sigma_t \tilde{\ell}(f, Z_{t})
    - \sigma_t \max_k \ell(f(\tilde{X_{t}} + \delta_k), \tilde{Y_{t}})
    + \sigma_t \max_k \ell(f(\tilde{X_{t}} + \delta_k), \tilde{Y_{t}})
\bigg] \\
\leq &\text{Lip}^1(\ell) \cdot \frac{a_1 \tau}{2}
+ \frac{1}{n} \sum_{t=1}^n \left[
    \sigma_t \max_k \ell(f(\tilde{X_{t}} + \delta_k), \tilde{Y_{t}})
\right]
\end{align*}

This leads to an upper bound of \(\widehat{\mathcal{R}}_n(\mathcal{L}_n)\) as follows:

\begin{align*}
\widehat{\mathcal{R}}_n(\mathcal{L}_n)
\leq\, \mathbb{E}_\sigma \bigg[
    \sup_{f \in \mathcal{NN}_{U, L}^{a_1, a_2} }
    \frac{1}{n} \sum_{t=1}^n \sigma_t \max_k \ell(f(\tilde{X_{t}} + \delta_k), \tilde{Y_{t}})
\bigg] 
+ \text{Lip}^1(\ell) \cdot \frac{a_1 \tau}{2}
\end{align*}
Define the class
\[
\mathcal{L}_{n,\tau} = \left\{ \max_k \ell(f(x + \delta_k), y) : \mathcal{Z} \mapsto \mathbb{R} \mid f \in \mathcal{NN}_{U, L}^{a_1, a_2}  \right\}.
\]

Let \(\mathcal{N}(u, \mathcal{L}_{n,\tau}, L_\infty(P_n))\) denote the covering number of the class \(\mathcal{L}_{n,\tau}\) under the data-dependent \(L_\infty\) metric. Define
\[
S_{n,M_\tau} = \{ x_i + \delta_k : i = 1, \ldots, n, \; K = 1, \ldots, M_\tau \},
\]
and let \(\mathcal{N}(u, \mathcal{NN}_{U, L}^{a_1, a_2} , L_\infty(P_{nM_\tau}))\) denote the covering number of the class \(\mathcal{NN}_{U, L}^{a_1, a_2} \) under the \(L_\infty\) metric on the set \(S_{n,M_\tau}\). For any \(f, f' \in \mathcal{NN}_{U, L}^{a_1, a_2} \), if \(\max_{t,k} |f(\tilde{X_{t}} + \delta_k) - f'(\tilde{X_{t}} + \delta_k)| \leq u\), then

\begin{align*}
\max_t \big|& \max_k \ell(f(\tilde{X_{t}} + \delta_k), \tilde{Y_{t}})
    - \max_k \ell(f'(\tilde{X_{t}} + \delta_k), \tilde{Y_{t}}) \big|\\
&\leq \max_{t,k} \big| \ell(f(\tilde{X_{t}} + \delta_k), \tilde{Y_{t}})
    - \ell(f'(\tilde{X_{t}} + \delta_k), \tilde{Y_{t}}) \big| \\
&\leq \text{Lip}^1(\ell) \cdot u
\end{align*}

Hence,
\[
\mathcal{N}(u, \mathcal{L}_{n,\tau}, L_\infty(P_n)) \leq \mathcal{N}(u / \text{Lip}^1(\ell), \mathcal{NN}_{U, L}^{a_1, a_2} , L_\infty(P_{nM_\tau})).
\]
Suppose the functions in \(\mathcal{NN}_{U, L}^{a_1, a_2} \) are uniformly bounded. Then the uniform covering number of \(\mathcal{NN}_{U, L}^{a_1, a_2} \) is defined by
\[
\mathcal{N}_\infty(u, \mathcal{NN}_{U, L}^{a_1, a_2} , nM_\tau) = \sup_{P_n} \mathcal{N}(u, \mathcal{NN}_{U, L}^{a_1, a_2} , L_\infty(P_nM_\tau)),
\]
where the supremum runs over all data sets of size \(n\). Combining Lemma~\ref{lem:5} and Lemma~\ref{lem:4}, we derive
\begin{align*}
\log \mathcal{N}(u, \mathcal{L}_{n,\tau},& L_\infty(P_n))  \leq \log \mathcal{N}_\infty(u, \mathcal{NN}_{U, L}^{a_1, a_2} , nM_\tau) \\
&\leq B_1 W^2 L^2 (L + \log(W^2 L)) \log(u^{-1}nM_\tau),
\end{align*}
for a constant \(B_1\).  Since \(\mathcal{NN}_{U, L}^{a_1, a_2} \) is bounded and \(\ell\) is continuous, there exists \(B_2 > 0\) such that
\[
\sup_{Z_{t} \in {P}_{\mathrm{mix}}} \max_k \ell(f(\tilde{X_{t}} + \delta_k), \tilde{Y_{t}}) \leq B_2 \quad \text{for any } f \in \mathcal{NN}_{U, L}^{a_1, a_2} .
\]
Combining above with Lemma~\ref{lem:2}, we have
\begin{align*}
\mathbb{E}_\sigma& \left\{ \sup_{f \in \mathcal{NN}_{U, L}^{a_1, a_2} } \frac{1}{n} \sum_{t=1}^n \sigma_t \max_k \ell(f(\tilde{X_{t}} + \delta_k), \tilde{Y_{t}}) \right\} \\
&\leq \inf_{\delta \geq 0} \left\{ 4\delta + 12 \int_{\delta}^{B_2} \sqrt{\frac{\log \mathcal{N}(u, \mathcal{L}_{n,\tau}, L_\infty(P_n))}{n}} \, du \right\}.
\end{align*}

Hence,
\begin{align*}
\lesssim \inf_{\delta \geq 0} \bigg[ \delta + \bigg( \sqrt{UL^2 + UL\log(U )}\bigg)n^{-1/2}  \cdot \int_{\delta}^{B_2} 
\left[ \sqrt{\log(u^{-1})} + \sqrt{\log n} + \sqrt{\log M_\tau} \right] \, du \bigg].
\end{align*}

Thus,
\[
\lesssim \bigg( \sqrt{UL^2 + UL\log(U )}\bigg) n^{-1/2} \left\{ \sqrt{\log n} + \sqrt{\log(\rho \tau^{-1})} \right\}.
\]

Therefore, by selecting \(\tau\) such that \(\rho \tau^{-1} = \mathcal{O}(n)\), we show
\begin{align*}
| D _{P_{\mathrm{mix}},\rho}(\hat f^{mix}_{NN}) -  &D^{mix}_{P_{n},\rho}(\hat f^{mix}_{NN})| \lesssim \rho n^{-1} +\bigg( \sqrt{U+ U\log(U )}\bigg)n^{-1/2} \sqrt{\log n}.
\end{align*}
Now if we summarize them all, we get:
\begin{align*}
| D _{P_{\mathrm{mix}},\rho}(\hat f^{mix}_{NN}) -   D _{P_{\mathrm{mix}},\rho}(f_*^{mix}) | 
&\lesssim \frac{\alpha^d}{Z^{\alpha} } +    \rho\,\sqrt{\frac{t}{2n}} + \rho^2 + n^{-\min\{1/2, \alpha/d\}} \log^{c(\alpha, d)} n + \rho n^{-1} \\&+\bigg(\sqrt{U + U\log(U)}\bigg)n^{-1/2}\sqrt{\log n}.
\end{align*}

\FloatBarrier
\section{Additional Method Details and Results}
\label{appendix:d}
\subsection{Additional Robustness and Calibration Results}
\begin{table}[htbp]
\centering
\caption{Average accuracy (\%) and ECE (\%) of WideResNet-28$\times$2 on CIFAR and CIFAR-C benchmarks. Higher accuracy is better and lower ECE is better.}
\vspace{1\baselineskip}
\resizebox{\textwidth}{!}{%
  \begin{tabular}{lcccccccc}
    \toprule
    \multirow{3}{*}{Method} &
      \multicolumn{4}{c}{CIFAR-10} &
      \multicolumn{4}{c}{CIFAR-100} \\
    \cmidrule(r){2-5}\cmidrule(l){6-9}
     & \multicolumn{2}{c}{Clean} & \multicolumn{2}{c}{-C} &
       \multicolumn{2}{c}{Clean} & \multicolumn{2}{c}{-C} \\
    \cmidrule(r){2-3}\cmidrule(r){4-5}\cmidrule(r){6-7}\cmidrule(l){8-9}
     & Acc. $\uparrow$ & ECE $\downarrow$ & Acc. $\uparrow$ & ECE $\downarrow$ &
       Acc. $\uparrow$ & ECE $\downarrow$ & Acc. $\uparrow$ & ECE $\downarrow$ \\
    \midrule
    Baseline        & 94.86 & 3.24 & 74.37 & 17.95 & 75.95 & 11.67 & 44.51 & 33.58 \\
    \midrule
    Mixup           & 94.72 & 20.95 & 77.96 & 16.69 & 75.17 & 16.58 & 48.72 & 13.05 \\
    Mixup + DRO     & 95.77 & 12.12 & 78.72 & 9.67 & 77.36 & 5.78 & 49.71 & 12.62 \\
    \midrule
    Manifold Mixup      & 95.81 & 12.12 & 75.38 & 11.41 & 77.79 & 8.36 & 48.84 & 13.12 \\
    Manifold Mixup + DRO & 95.52 & 7.76 & 75.65 & 9.65 & 77.86 & 5.03 & 48.95 & 12.01 \\
    \midrule
    NoisyMix        & 95.31 & 10.67 & 90.50 & 10.97 & 75.70 & 26.75 & 65.67 & 24.59 \\
    NoisyMix + DRO  & 95.36 & 7.12 & 90.18 & 6.46 & 76.00 & 20.00 & 65.87 & 18.44 \\
    \bottomrule
  \end{tabular}}
\label{tab:wideresnet28x2-augmentation-comparison}
\end{table}

\begin{table*}[htbp]
\centering
\caption{Sensitivity study of the DRO radius $\rho$ for Mixup + DRO, NoisyMix + DRO, and Manifold Mixup + DRO on CIFAR and CIFAR-C benchmarks using PreActResNet-18. Metrics follow Table~\ref{tab:augmentation-comparison}: higher accuracy is better and lower ECE is better.}
\vspace{1\baselineskip}
\resizebox{\textwidth}{!}{%
  \begin{tabular}{llcccccccc}
    \toprule
    \multirow{3}{*}{Method} & \multirow{3}{*}{$\rho$} &
      \multicolumn{4}{c}{CIFAR‑10} &
      \multicolumn{4}{c}{CIFAR‑100} \\
    \cmidrule(r){3-6}\cmidrule(l){7-10}
     & & \multicolumn{2}{c}{Clean} & \multicolumn{2}{c}{‑C} &
       \multicolumn{2}{c}{Clean} & \multicolumn{2}{c}{‑C} \\
    \cmidrule(r){3-4}\cmidrule(r){5-6}\cmidrule(r){7-8}\cmidrule(l){9-10}
     & & Acc. $\uparrow$ & ECE $\downarrow$ & Acc. $\uparrow$ & ECE $\downarrow$ &
       Acc. $\uparrow$ & ECE $\downarrow$ & Acc. $\uparrow$ & ECE $\downarrow$ \\
    \midrule
    \multirow{6}{*}{Mixup + DRO}
      & $1.0 \times 10^{-4}$ & 96.06 & 6.30 & 80.08 & 7.20 & 80.15 & 5.77 & 54.41 & 9.24 \\
      & $2.0 \times 10^{-4}$ & 96.09 & 6.03 & 79.52 & 6.84 & 80.17 & 7.57 & 54.43 & 10.54 \\
      & $3.0 \times 10^{-4}$ & 95.86 & 6.46 & 79.74 & 7.92 & 80.38 & 9.81 & 54.44 & 9.73 \\
      & $5.0 \times 10^{-4}$ & 96.13 & 6.18 & 80.04 & 6.79 & 80.07 & 8.08 & 54.31 & 9.13 \\
      & $2.0 \times 10^{-3}$ & 96.04 & 7.54 & 80.75 & 7.25 & 80.25 & 8.95 & 54.24 & 9.80 \\
      & $5.0 \times 10^{-3}$ & 96.24 & 5.81 & 79.76 & 7.22 & 80.47 & 8.25 & 54.88 & 9.85 \\
    \midrule
    \multirow{6}{*}{NoisyMix + DRO}
      & $1.0 \times 10^{-4}$ & 95.56 & 8.35 & 91.80 & 7.60 & 79.99 & 14.93 & 71.42 & 12.90 \\
      & $2.0 \times 10^{-4}$ & 95.60 & 8.74 & 91.73 & 7.90 & 80.38 & 15.27 & 71.21 & 13.02 \\
      & $3.0 \times 10^{-4}$ & 95.73 & 8.49 & 91.72 & 7.77 & 80.44 & 15.15 & 71.46 & 12.95 \\
      & $5.0 \times 10^{-4}$ & 95.66 & 7.26 & 91.57 & 6.97 & 80.13 & 14.29 & 71.15 & 12.20 \\
      & $2.0 \times 10^{-3}$ & 95.56 & 7.65 & 91.49 & 7.26 & 80.10 & 14.26 & 71.18 & 12.09 \\
      & $5.0 \times 10^{-3}$ & 95.53 & 8.05 & 91.47 & 7.48 & 80.08 & 15.13 & 71.12 & 12.99 \\
    \midrule
    \multirow{6}{*}{Manifold Mixup + DRO}
      & $1.0 \times 10^{-4}$ & 95.86 & 9.87 & 77.18 & 11.75 & 80.25 & 9.55 & 53.65 & 12.13 \\
      & $2.0 \times 10^{-4}$ & 95.74 & 9.83 & 77.44 & 10.85 & 79.98 & 8.81 & 54.25 & 11.77 \\
      & $3.0 \times 10^{-4}$ & 95.75 & 9.71 & 76.91 & 10.66 & 80.52 & 9.42 & 53.26 & 12.45 \\
      & $5.0 \times 10^{-4}$ & 95.86 & 9.13 & 77.24 & 10.62 & 79.98 & 9.55 & 54.09 & 11.84 \\
      & $2.0 \times 10^{-3}$ & 95.73 & 9.40 & 78.12 & 10.10 & 80.25 & 9.10 & 53.61 & 13.84 \\
      & $5.0 \times 10^{-3}$ & 95.86 & 9.44 & 77.99 & 10.68 & 80.56 & 9.57 & 54.35 & 11.30 \\
    \bottomrule
  \end{tabular}}
\label{tab:rho-sensitivity-mixup-noisymix}
\end{table*}

\begin{table*}[htbp]
\centering
\caption{Sensitivity study of the mixing-strength parameter $\alpha_{\mathrm{mix}}$ on CIFAR and CIFAR-C benchmarks using PreActResNet-18. Higher accuracy is better and lower ECE is better. Bracketed values report the corresponding method with DRO.}
\vspace{1\baselineskip}
\resizebox{\textwidth}{!}{%
  \begin{tabular}{llcccccccc}
    \toprule
    \multirow{3}{*}{Method} & \multirow{3}{*}{$\alpha_{\mathrm{mix}}$} &
      \multicolumn{4}{c}{CIFAR‑10} &
      \multicolumn{4}{c}{CIFAR‑100} \\
    \cmidrule(r){3-6}\cmidrule(l){7-10}
     & & \multicolumn{2}{c}{Clean} & \multicolumn{2}{c}{‑C} &
       \multicolumn{2}{c}{Clean} & \multicolumn{2}{c}{‑C} \\
    \cmidrule(r){3-4}\cmidrule(r){5-6}\cmidrule(r){7-8}\cmidrule(l){9-10}
     & & Acc. $\uparrow$ & ECE $\downarrow$ & Acc. $\uparrow$ & ECE $\downarrow$ &
       Acc. $\uparrow$ & ECE $\downarrow$ & Acc. $\uparrow$ & ECE $\downarrow$ \\
    \midrule
    \multirow{4}{*}{Mixup [+DRO]}
      & 0.2 & 95.69 [95.75] & 2.52 [1.12] & 76.97 [77.99] & 10.14 [10.98] & 79.02 [79.33] & 3.53 [1.40] & 52.17 [51.96] & 8.46 [10.69] \\
      & 0.4 & 95.84 [95.97] & 11.38 [2.88] & 78.45 [79.46] & 10.89 [7.87] & 78.66 [79.62] & 12.62 [6.00] & 51.71 [52.15] & 10.90 [9.37] \\
      & 0.6 & 95.60 [96.15] & 11.32 [4.12] & 78.66 [80.29] & 10.44 [6.56] & 79.54 [80.02] & 12.80 [7.28] & 53.03 [53.44] & 10.16 [8.85] \\
      & 1.0 & 95.69 [96.01] & 14.70 [5.95] & 79.16 [79.69] & 12.57 [7.47] & 79.69 [80.31] & 14.96 [8.32] & 52.83 [55.00] & 11.69 [9.21] \\
    \midrule
    \multirow{4}{*}{NoisyMix [+DRO]}
      & 0.2 & 95.59 [95.42] & 4.48 [1.74] & 91.49 [91.25] & 4.17 [2.65] & 76.70 [77.27] & 5.01 [1.99] & 69.04 [69.08] & 4.07 [5.74] \\
      & 0.4 & 95.57 [95.59] & 10.05 [8.71] & 91.25 [91.28] & 9.29 [7.80] & 78.22 [78.47] & 13.57 [11.36] & 70.22 [70.46] & 11.94 [9.20] \\
      & 0.6 & 95.78 [95.70] & 10.27 [3.45] & 90.81 [90.08] & 9.22 [4.50] & 79.12 [79.52] & 12.70 [6.11] & 70.22 [69.88] & 10.53 [4.89] \\
      & 1.0 & 95.27 [95.60] & 20.39 [11.08] & 91.45 [91.59] & 21.17 [10.72] & 79.58 [79.85] & 23.59 [16.17] & 71.26 [71.20] & 21.67 [14.03] \\
    \midrule
    \multirow{4}{*}{Manifold Mixup [+DRO]}
      & 0.4 & 95.84 [95.97] & 13.90 [4.23] & 77.96 [78.53] & 11.54 [9.13] & 80.08 [79.99] & 8.00 [3.36] & 52.25 [52.21] & 10.50 [10.91] \\
      & 0.8 & 95.83 [95.95] & 14.59 [6.00] & 77.64 [76.82] & 12.89 [11.23] & 79.75 [80.50] & 8.28 [6.13] & 53.25 [52.93] & 11.26 [11.10] \\
      & 1.2 & 95.86 [95.73] & 16.96 [8.80] & 78.38 [76.70] & 14.01 [10.98] & 79.45 [80.21] & 12.36 [8.96] & 52.54 [53.70] & 12.78 [11.96] \\
      & 2.0 & 95.83 [95.85] & 17.44 [10.09] & 78.71 [77.66] & 15.11 [10.38] & 80.19 [80.20] & 12.83 [9.48] & 53.31 [53.19] & 12.73 [12.17] \\
    \bottomrule
  \end{tabular}}
\label{tab:mixup-alpha-ablation-cifar}
\end{table*}

\FloatBarrier
\subsection{Dataset Details}

The CIFAR-10-C and CIFAR-100-C datasets \citep{hendrycks2019benchmarking} are widely used to evaluate model robustness under synthetic/natural corruptions.
These datasets are created by introducing 15 distinct types of corruption (including, but not limited to, noise, blur, weather effects, and digital distortions) to the CIFAR-10 and CIFAR-100 datasets, respectively.
Each of these fifteen corruption types is applied at five predefined severity levels, which control the intensity of the corruption based on human perceptual assessment.
These corrupted datasets enable a systematic analysis of model performance at various levels and types of corruption.

\FloatBarrier
\subsection{Training Details}

For the experimental setup, we train our models using stochastic gradient descent (SGD) with Nesterov momentum. The momentum coefficient is set to 0.9, and the weight decay is fixed at 0.0005. The initial learning rate is set to $1 \times 10^{-1}$ and follows a cosine annealing schedule, gradually decaying to a minimum value of $1 \times 10^{-5}$ over the course of training. We use a mini-batch size of 128 for training and a batch size of 1000 during evaluation. For Mixup, Manifold Mixup, and NoisyMix, we adopt the recommended hyperparameter values as specified in their original papers.

For the main experiments, we use $\alpha_{\mathrm{mix}}=1$ for Mixup and NoisyMix and $\alpha_{\mathrm{mix}}=2$ for Manifold Mixup, following the settings used in their original papers.

For the PreActResNet-18 DRO-Augment results in Table~\ref{tab:augmentation-comparison}, we use $\rho=2.0\times10^{-3}$ for Mixup, $\rho=5.0\times10^{-3}$ for Manifold Mixup, and $\rho=3.0\times10^{-4}$ for NoisyMix.

All reported training runs use random seed 1024.

The bracketed +DRO results in Table~\ref{tab:mixup-alpha-ablation-cifar} use a fixed $\rho=1.0\times10^{-4}$. Unless otherwise noted, experiments are conducted on local GPU workstations equipped with two NVIDIA RTX A6000 GPUs (48 GB each). The $\alpha_{\mathrm{mix}}$ sensitivity experiments in Table~\ref{tab:mixup-alpha-ablation-cifar} were conducted separately on NVIDIA H200 GPUs. For CIFAR-10-C and CIFAR-100-C, the average training time per epoch under the DRO-Augment setting is approximately 0.019 hours.
The full implementation and all experiment scripts are publicly available at
https://anonymous.4open.science/r/DRO-Augment-6F2F.

\FloatBarrier
\subsection{DRO-Augment Training Procedure}

\begin{algorithm}[H]
\caption{Training with DRO-Augment}
\label{alg:custom_loss_augment}
\begin{algorithmic}
\Require Training data $\mathcal{D}$, neural network model $f_{\theta}$, objective function $\mathcal{L}$, augmentation function $\mathcal{A}$, training epochs $N$, learning rate $\eta$, regularization weight $\rho$.

\State Initialize model parameters $\theta$;
\For{epoch $= 1$ to $N$}
    \For{each minibatch $\mathcal{B} \subset \mathcal{D}$}
        \State Apply data augmentation: $\tilde{\mathcal{B}} = \mathcal{A}(\mathcal{B})$;
        \For{each input $(x_i, y_i) \in \tilde{\mathcal{B}}$}
            \State Compute model output: $\hat{y}_i = f_{\theta}(x_i)$;
            \State Compute total loss:
            \begin{equation*}
            \textstyle
            \mathcal{L}_{total} = \sum_i\mathcal{L}(f_{\theta}(x_i), y_i) + \rho \Bigl( \mathbb{E}_{P_n} \left[ \|\nabla_x \mathcal{L}(f_{\theta}(x_i), y_i) \|^q_*  \right]\Bigr)^{\!\!1/q};
            \end{equation*}
        \EndFor
        \State Update $\theta$ using stochastic gradient descent:
        \[
        \theta \gets \theta - \eta \nabla_{\theta} \mathcal{L}_{total};
        \]
    \EndFor
\EndFor
\end{algorithmic}
\end{algorithm}

\FloatBarrier
\FloatBarrier
\section{Refined CIFAR-C}
\label{appendix:e}
We aim to address the issue with the severity settings in the CIFAR-C datasets. In the original CIFAR-10-C dataset, model performance varies significantly across different corruption types, even at the same severity level. As the severity increases, this performance gap widens, with differences in accuracy exceeding 40\% in extreme cases. Additionally, at severity level 1—the lowest level—most baseline models, such as ResNet-18, already achieve near-maximum accuracy, around 95\%. This leaves minimal room for observing improvements, rendering severity level 1 practically uninformative.

To address this, we propose redefining the severity levels. Consistent with the architectural scope of this work, we design the refined
severity levels using ResNet performance as the reference. We set the baseline accuracy at 50\%, corresponding to the midpoint of the binary correct-versus-incorrect accuracy scale. For the upper bound, we set the initial accuracy at approximately 85\%, acknowledging that 95\% represents a practical ceiling for CIFAR-10-C. This adjustment provides sufficient room to observe robustness improvements across a broader range of models.

Table \ref{tab:refined-cifar10c} and \ref{tab:refined-cifar100c} show that, in CIFAR-100-C, ResNet performance generally drops by approximately 30\% compared to CIFAR-10-C. To account for this, we adjust our standard accuracy range downward by 30\%. Specifically, we redefine the severity levels such that at severity 1, the accuracy is set at 60\%, decreasing in increments of 10\% down to 20\% at severity 5. This proposed configuration ensures more consistent accuracy across different corruption types at the same severity level for ResNet architectures of varying depths, allowing for a more reliable evaluation of model robustness. The detailed performance of ResNet-18, ResNet-50, and ResNet-101 on the refined CIFAR-10-C and CIFAR-100-C benchmarks can be found in table~\ref{tab:resnet18-cifar10c}, \ref{tab:resnet50-cifar10c-third}, \ref{tab:resnet101-cifar10c-fourth}, \ref{tab:resnet18-cifar100c-refined}, \ref{tab:resnet50-cifar100c-refined}, and \ref{tab:resnet101-cifar100c-refined}.
\begin{table}[htbp]
\centering
\setlength{\tabcolsep}{5pt}
\caption{Performance on Refined CIFAR\textendash10\textendash C (\% accuracy)}
\vspace{1\baselineskip} 
\begin{tabular}{@{}lccccc@{}}
\toprule
\textbf{Model} & S1 & S2 & S3 & S4 & S5 \\
\midrule
ResNet-18 & 86.04 & 79.03 & 70.00 & 61.83 & 51.59 \\
ResNet-50 & 88.10 & 80.18 & 70.18 & 61.69 & 50.68 \\
ResNet-101 & 87.60 & 79.54 & 69.51 & 60.85 & 49.74 \\
\bottomrule
\end{tabular}
\label{tab:refined-cifar10c}
\vspace{-0.5em} 
\end{table}

\begin{table}[htbp]
\centering
\setlength{\tabcolsep}{5pt}
\caption{Performance on Refined CIFAR\textendash100\textendash C (\% accuracy)}
\vspace{1\baselineskip} 
\begin{tabular}{@{}lccccc@{}}
\toprule
\textbf{Model} & S1 & S2 & S3 & S4 & S5 \\
\midrule
ResNet-18 & 59.93 & 50.78 & 40.85 & 30.77 & 20.53 \\
ResNet-50 & 65.11 & 55.83 & 45.51 & 35.05 & 23.86 \\
ResNet-101 & 63.80 & 54.58 & 44.65 & 34.53 & 23.87 \\

\bottomrule
\end{tabular}
\label{tab:refined-cifar100c}
\vspace{-0.5em} 
\end{table}

\begin{table}[h]
\centering
\caption{Accuracy of ResNet-18 on Refined CIFAR-10-C under different severity levels}
\vspace{1\baselineskip} 
\begin{tabular}{lccccc}
\toprule
\textbf{Type} & \textbf{S1} & \textbf{S2} & \textbf{S3} & \textbf{S4} & \textbf{S5} \\
\midrule
gaussian   & 86.20 & 78.01 & 69.30 & 61.61 & 51.22 \\
shot       & 86.51 & 77.16 & 69.00 & 60.36 & 50.88 \\
impulse    & 85.01 & 78.65 & 69.19 & 59.73 & 50.82 \\
defocus    & 87.03 & 79.53 & 69.27 & 60.50 & 46.63 \\
glass      & 84.88 & 79.16 & 67.59 & 60.73 & 52.56 \\
motion     & 85.30 & 78.56 & 70.38 & 60.05 & 52.34 \\
zoom       & 85.81 & 79.94 & 71.49 & 64.12 & 54.39 \\
snow       & 87.02 & 78.12 & 69.72 & 64.15 & 54.69 \\
frost      & 85.86 & 77.90 & 69.68 & 63.13 & 54.17 \\
fog        & 85.38 & 78.62 & 69.98 & 60.35 & 49.59 \\
bright     & 86.83 & 80.06 & 71.70 & 62.78 & 52.40 \\
contrast   & 86.62 & 81.31 & 71.98 & 63.79 & 52.49 \\
elastic    & 85.55 & 78.86 & 69.09 & 59.94 & 51.29 \\
pixelate   & 86.49 & 80.47 & 71.57 & 64.92 & 53.53 \\
jpeg       & 86.18 & 79.03 & 70.01 & 61.33 & 46.79 \\
\midrule
\textbf{Avg} & \textbf{86.04} & \textbf{79.03} & \textbf{70.00} & \textbf{61.83} & \textbf{51.59} \\
\bottomrule
\end{tabular}
\label{tab:resnet18-cifar10c}
\vspace{-1em}
\end{table}

\begin{table}[h]
\centering
\caption{Accuracy of ResNet-50 on Refined CIFAR-10-C under different severity levels}
\vspace{1\baselineskip} 
\begin{tabular}{lccccc}
\toprule
\textbf{Type} & \textbf{S1} & \textbf{S2} & \textbf{S3} & \textbf{S4} & \textbf{S5} \\
\midrule
gaussian   & 88.40 & 80.30 & 70.74 & 62.44 & 50.43 \\
shot       & 89.17 & 79.85 & 71.92 & 62.72 & 51.95 \\
impulse    & 86.97 & 81.18 & 70.95 & 61.20 & 50.78 \\
defocus    & 89.28 & 78.65 & 64.31 & 53.83 & 39.72 \\
glass      & 86.69 & 80.06 & 67.51 & 61.80 & 52.19 \\
motion     & 87.08 & 78.72 & 69.60 & 59.01 & 49.99 \\
zoom       & 87.74 & 79.16 & 68.41 & 59.56 & 48.26 \\
snow       & 88.91 & 79.66 & 69.56 & 65.10 & 53.84 \\
frost      & 88.10 & 79.13 & 72.06 & 63.32 & 54.37 \\
fog        & 87.65 & 81.19 & 72.50 & 64.43 & 52.57 \\
bright     & 88.92 & 81.53 & 71.93 & 62.97 & 52.53 \\
contrast   & 88.85 & 83.61 & 75.32 & 67.76 & 57.22 \\
elastic    & 86.89 & 79.88 & 70.57 & 60.93 & 51.27 \\
pixelate   & 88.32 & 79.58 & 68.12 & 60.87 & 49.69 \\
jpeg       & 88.53 & 80.15 & 69.14 & 59.44 & 45.43 \\
\midrule
\textbf{Avg} & \textbf{88.10} & \textbf{80.18} & \textbf{70.18} & \textbf{61.69} & \textbf{50.68} \\
\bottomrule
\end{tabular}
\label{tab:resnet50-cifar10c-third}
\end{table}

\begin{table}[h]
\centering
\caption{Accuracy of ResNet-101 on Refined CIFAR-10-C  under different severity levels}
\vspace{1\baselineskip} 
\begin{tabular}{lccccc}
\toprule
\textbf{Type} & \textbf{S1} & \textbf{S2} & \textbf{S3} & \textbf{S4} & \textbf{S5} \\
\midrule
gaussian   & 87.93 & 79.14 & 70.12 & 61.87 & 51.49 \\
shot       & 88.16 & 78.44 & 69.57 & 61.25 & 51.38 \\
impulse    & 86.71 & 80.30 & 70.53 & 60.33 & 51.44 \\
defocus    & 88.89 & 78.76 & 65.34 & 54.08 & 38.82 \\
glass      & 86.35 & 79.18 & 67.53 & 63.04 & 51.37 \\
motion     & 86.86 & 78.19 & 69.17 & 57.63 & 48.23 \\
zoom       & 87.30 & 79.16 & 68.50 & 60.03 & 48.15 \\
snow       & 87.98 & 78.10 & 68.25 & 63.24 & 50.69 \\
frost      & 87.43 & 77.59 & 68.63 & 60.51 & 51.59 \\
fog        & 87.01 & 80.17 & 70.99 & 61.46 & 49.88 \\
bright     & 88.28 & 80.47 & 69.92 & 60.25 & 50.06 \\
contrast   & 88.22 & 82.30 & 73.23 & 65.31 & 54.19 \\
elastic    & 86.73 & 79.79 & 69.99 & 60.35 & 51.00 \\
pixelate   & 88.19 & 81.69 & 70.86 & 63.70 & 51.89 \\
jpeg       & 87.94 & 79.82 & 70.00 & 59.66 & 45.87 \\
\midrule
\textbf{Avg} & \textbf{87.60} & \textbf{79.54} & \textbf{69.51} & \textbf{60.85} & \textbf{49.74} \\
\bottomrule
\end{tabular}
\label{tab:resnet101-cifar10c-fourth}
\end{table}

\begin{table}[h]
\centering
\caption{Accuracy of ResNet-18 on Refined CIFAR-100-C under different severity levels}
\vspace{1\baselineskip} 
\begin{tabular}{lccccc}
\toprule
\textbf{Type} & \textbf{S1} & \textbf{S2} & \textbf{S3} & \textbf{S4} & \textbf{S5} \\
\midrule
gaussian   & 61.88 & 49.74 & 39.15 & 31.80 & 19.09 \\
shot       & 61.45 & 50.45 & 42.01 & 29.19 & 18.49 \\
impulse    & 61.29 & 52.34 & 42.23 & 29.00 & 22.49 \\
defocus    & 61.01 & 50.95 & 40.08 & 33.01 & 21.55 \\
glass      & 59.79 & 50.35 & 38.33 & 25.36 & 19.48 \\
motion     & 59.86 & 52.02 & 41.66 & 32.28 & 20.51 \\
zoom       & 57.44 & 48.62 & 40.52 & 29.88 & 18.71 \\
snow       & 59.39 & 53.55 & 43.97 & 32.79 & 21.91 \\
frost      & 58.51 & 48.81 & 42.59 & 30.20 & 19.89 \\
fog        & 58.13 & 52.68 & 40.17 & 31.70 & 23.20 \\
bright     & 58.87 & 52.06 & 41.42 & 29.47 & 20.36 \\
contrast   & 60.28 & 51.64 & 39.66 & 32.77 & 21.74 \\
elastic    & 59.93 & 49.45 & 40.41 & 32.73 & 19.08 \\
pixelate   & 61.13 & 49.21 & 40.00 & 31.69 & 19.89 \\
jpeg       & 59.99 & 49.76 & 40.59 & 29.75 & 21.60 \\
\midrule
\textbf{Avg} & \textbf{59.93} & \textbf{50.78} & \textbf{40.85} & \textbf{30.77} & \textbf{20.53} \\
\bottomrule
\end{tabular}
\label{tab:resnet18-cifar100c-refined}
\end{table}

\begin{table}[h]
\centering
\caption{Accuracy of ResNet-50 on Refined CIFAR-100-C under different severity levels}
\vspace{1\baselineskip} 
\begin{tabular}{lccccc}
\toprule
\textbf{Type} & \textbf{S1} & \textbf{S2} & \textbf{S3} & \textbf{S4} & \textbf{S5} \\
\midrule
gaussian   & 66.76 & 54.96 & 44.10 & 36.41 & 22.75 \\
shot       & 66.23 & 56.02 & 47.55 & 34.17 & 22.47 \\
impulse    & 66.29 & 59.35 & 48.46 & 35.49 & 28.00 \\
defocus    & 66.02 & 55.38 & 42.41 & 35.78 & 23.08 \\
glass      & 65.20 & 53.53 & 40.20 & 27.68 & 21.45 \\
motion     & 64.95 & 55.80 & 44.60 & 34.38 & 20.64 \\
zoom       & 62.35 & 52.88 & 43.74 & 32.22 & 19.69 \\
snow       & 65.11 & 59.60 & 51.18 & 39.26 & 27.21 \\
frost      & 64.19 & 54.81 & 48.72 & 36.35 & 25.50 \\
fog        & 63.85 & 58.84 & 45.78 & 37.20 & 27.33 \\
bright     & 64.84 & 58.08 & 48.39 & 36.03 & 25.60 \\
contrast   & 65.36 & 58.06 & 47.06 & 40.63 & 28.80 \\
elastic    & 64.71 & 53.68 & 44.66 & 36.46 & 21.55 \\
pixelate   & 66.05 & 52.94 & 43.46 & 33.07 & 22.21 \\
jpeg       & 64.71 & 53.56 & 42.33 & 30.69 & 21.55 \\
\midrule
\textbf{Avg} & \textbf{65.11} & \textbf{55.83} & \textbf{45.51} & \textbf{35.05} & \textbf{23.86} \\
\bottomrule
\end{tabular}
\label{tab:resnet50-cifar100c-refined}
\end{table}

\begin{table}[h]
\centering
\caption{Accuracy of ResNet-101 on Refined CIFAR-100-C under different severity levels}
\vspace{1\baselineskip} 
\begin{tabular}{lccccc}
\toprule
\textbf{Type} & \textbf{S1} & \textbf{S2} & \textbf{S3} & \textbf{S4} & \textbf{S5} \\
\midrule
gaussian   & 65.58 & 54.04 & 44.05 & 36.71 & 24.64 \\
shot       & 64.83 & 54.97 & 47.32 & 35.17 & 24.26 \\
impulse    & 65.32 & 57.01 & 47.40 & 35.29 & 28.20 \\
defocus    & 65.11 & 54.15 & 42.78 & 36.00 & 23.80 \\
glass      & 64.06 & 54.62 & 42.62 & 30.01 & 23.17 \\
motion     & 64.00 & 54.06 & 43.82 & 34.58 & 21.41 \\
zoom       & 61.09 & 51.24 & 42.87 & 31.93 & 20.06 \\
snow       & 63.40 & 57.36 & 47.79 & 35.80 & 24.41 \\
frost      & 62.37 & 52.55 & 46.55 & 34.92 & 23.44 \\
fog        & 62.12 & 57.15 & 44.42 & 36.25 & 26.80 \\
bright     & 62.94 & 56.35 & 46.72 & 34.40 & 25.53 \\
contrast   & 63.93 & 56.04 & 44.92 & 39.09 & 27.84 \\
elastic    & 63.53 & 53.33 & 43.38 & 35.26 & 21.50 \\
pixelate   & 64.91 & 52.21 & 42.70 & 31.52 & 20.89 \\
jpeg       & 63.81 & 53.60 & 42.42 & 31.03 & 22.07 \\
\midrule
\textbf{Avg} & \textbf{63.80} & \textbf{54.58} & \textbf{44.65} & \textbf{34.53} & \textbf{23.87} \\
\bottomrule
\end{tabular}
\label{tab:resnet101-cifar100c-refined}
\end{table}

\FloatBarrier
\section{Use of LLMs}
LLM is used to aid or polish writing.

\end{document}